\documentclass[fleqn,10pt]{wlscirep}

\usepackage[utf8]{inputenc}
\usepackage[T1]{fontenc}

\title{An aerial color image anomaly dataset for search missions in complex forested terrain}
\usepackage[sorting=none]{biblatex}
\usepackage{array}

\newcolumntype{C}[1]{>{\centering\arraybackslash}p{#1}}

\DeclareFieldFormat[article]{volume}{\textbf{#1}}
\usepackage{makecell}
\usepackage{float}
\usepackage{hyperref}

\author[1,+]{Rakesh John Amala Arokia Nathan}
\author[2,+]{Matthias Gessner}
\author[1,+]{Nurullah Özkan}
\author[2]{Marius Bock}
\author[1]{Mohamed Youssef}
\author[2]{Maximilian Mews}
\author[2]{Björn Piltz}
\author[2]{Ralf Berger}
\author[1,*,+]{Oliver Bimber}
\affil[1]{Johannes Kepler University, Institute of Computer Graphics, Linz, 4040, Austria}
\affil[2]{German Aerospace Center, Institute of Space Research, Berlin, 12489, Germany}

\affil[*]{oliver.bimber@jku.at}

\affil[+]{these authors contributed equally to this work}

\begin{abstract}
After a family murder in rural Germany, authorities failed to locate the suspect in a vast forest despite a massive search. To aid the search, a research aircraft captured high-resolution aerial imagery. Due to dense vegetation obscuring small clues, automated analysis was ineffective, prompting a crowd-search initiative. This effort produced a unique dataset of labeled, hard-to-detect anomalies under occluded, real-world conditions. It can serve as a benchmark for improving anomaly detection approaches in complex forest environments, supporting manhunts and rescue operations. 
Initial benchmark tests showed existing methods performed poorly, highlighting the need for context-aware approaches. The  dataset is openly accessible for offline processing. An additional interactive web interface supports online viewing and dynamic growth by allowing users to annotate and submit new findings.
\end{abstract}
\begin{document}

\flushbottom
\maketitle
% * <john.hammersley@gmail.com> 2015-02-09T12:07:31.197Z:
%
%  Click the title above to edit the author information and abstract
%
\thispagestyle{empty}

%https://www.nature.com/sdata/submission-guidelines

%-----------------------------------------------
\section*{Background \& Summary}
%For Data Descriptors, this section should provide an overview of dataset, including the motivation for creating it, as well as outlining the potential reuse value where applicable. Any previous publications that used these data, in whole or in part, should be cited and briefly summarised. Introductions for Articles and Comments should provide a similar explanation of why the work was performed and what value you believe it adds. Please note that because Data Descriptors do not present results or analyses there is no formal mandate to cite a given amount of prior art for comparison, however we recommend that at least some other datasets or outputs relevant to the field are cited for readers' general interest. Please do not include subjective claims on novelty, impact, or utility. 

\begin{figure}
\centering
\includegraphics[width=\linewidth]{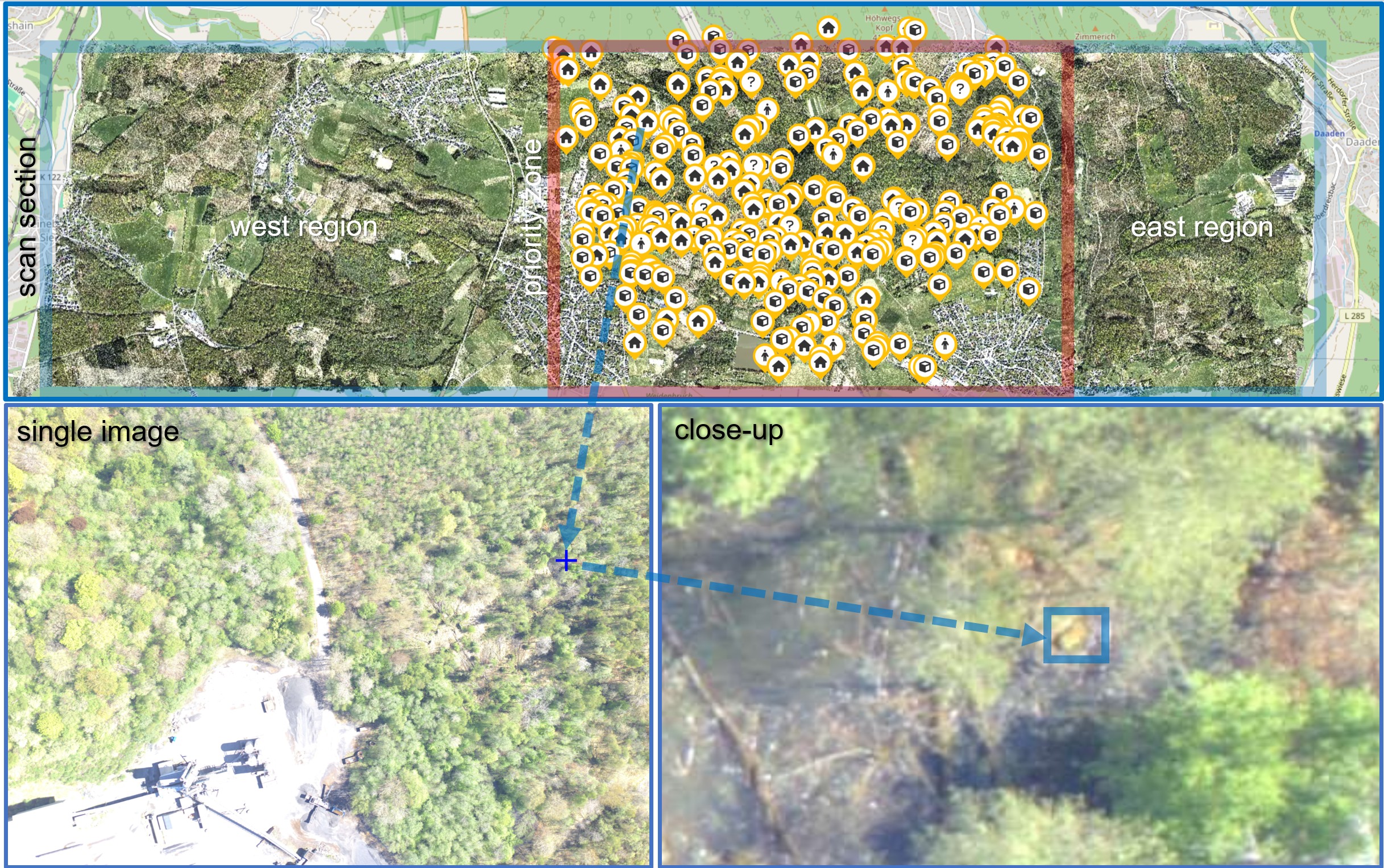}
\caption{Map of scan section and priority zone with orthographically projected aerial images of the flight and locations of relevant findings (top). Example finding in full aerial image and close-up (bottom). During ground observation, this finding was later identified as "a small barrel, very old, partially overgrown and filled with soil".}
\label{fig:map}
\end{figure}

\begin{figure}
\centering
\includegraphics[width=1\linewidth]{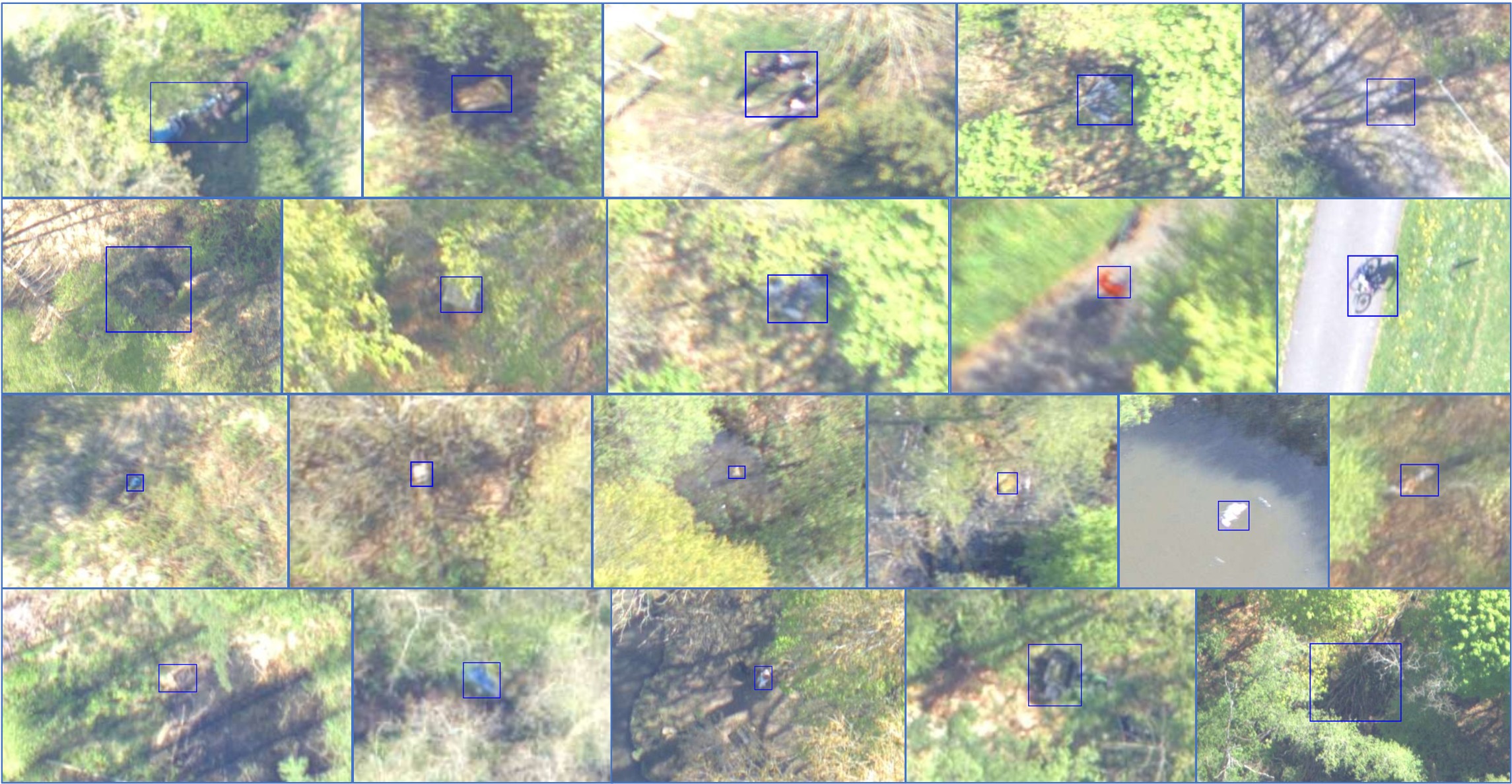}
\caption{Close-ups of various sample findings contained in the dataset together with their bounding-box labels. Overall, they include objects such as barrels, trash bags, tarps, metal barriers, floating objects on water; but also people and potential man-made shelters, hunting stands, sheds, huts, tents, shooting ranges, fire pits, and others -- all identified based on color and structural anomalies.}
\label{fig:example_closeups}
\end{figure}

On 6 April 2025, a family was brutally murdered in their home in the middle of the night in Weitefeld, a small German village. The prime suspect was believed to have fled into the nearby forest, a sprawling 60-square-kilometer area. Despite an extensive search involving over a thousand police officers, helicopters, drones, and divers, the perpetrator remained at large. The sheer size of the search area, combined with dense vegetation, made the operation extremely challenging. After three weeks with no breakthroughs, authorities concluded that the suspect was most likely dead. Yet, locating his remains was crucial—not only to bring closure to the case but also to ease the lingering fear among residents of Weitefeld and neighboring villages, who were left in the unsettling limbo of a cold case. To advance the search, Johannes Kepler University, the German Aerospace Center, and the Rhineland-Palatinate police. On 27 April 2025, a research aircraft took off from Würselen Airfield in Aachen, Germany, equipped with a specialized modular aerial camera system. Its mission was to scan a 25-square-kilometer section within the search area, capturing high-resolution (8,416x6,032 pixels) aerial imagery to detect anomalies, such as clothing or shelters that might indicate human remains. During the flight, the plane captured 30,454 RGB images with a  ground resolution of approximately 4x4 centimeters per pixel. However, due to dense vegetation obscuring potential clues —some of which might span just a few pixels— automated object classification was unrealistic. Instead, an online crowd-search initiative, enlisting 160 volunteers, was applied to a 10-square-kilometer priority zone of the scan section to manually analyze a subset of 10,659 images for irregularities. This crowd search was supported by automatic color anomaly detection \cite{reed1990adaptive}. A total of 405 anomalies were found and categorized as potential objects, shelters, persons, or unknown. 238 of these findings were considered relevant for the case and were subsequently verified by ground police teams (cf. Figure \ref{fig:map}). Despite these efforts, the mission ultimately did not locate the suspect, leaving the case unresolved. Nevertheless, the operation yielded a valuable dataset of 34,424 labeled and categorized anomalies across 405 findings appearing from different viewing angles and at different occlusion conditions in 10,659 aerial images (Figure \ref{fig:example_closeups} illustrates examples). Each of the findings was identified during ground observations by police, and the corresponding protocols are provided. Furthermore, we supply additional 19,795 unlabeled aerial images covering two non-priority zones (regions west and east of the priority zone) of the scan section, which facilitate future labeling and data acquisition. The dataset and the web interfaces used for the crowd-search is available online, enabling to view, analyze, and additionally label these images and findings. For validation, we applied our dataset to benchmark several common color anomaly detection methods, including deep-learning-based approaches (Feature Reconstruction Error (FRE) \cite{ndiour2022fre}, FastFlow \cite{yu2021fastflow}, Efficient AD \cite{batzner2024efficientad}) and model-based techniques (Reed–Xiaoli Global (RXG) \cite{reed1990adaptive}, Reed–Xiaoli Modified (RXM) \cite{chang2002anomaly}, principal component analysis (PCA) \cite{shyu2003novel}). We found that all of them performed poorly on our data -- especially in case of strong occlusion caused by vegetation. A key reason for this failure is that decisions based on local image regions or individual pixels lack the broader contextual information that our human volunteers utilized. For instance, occurring structures such as tree stumps, that are conspicuous in isolation may not be flagged as anomalies when appearing frequently across a wider area. We believe that our real-world, large-scale search effort under challenging and realistic conditions provides a unique resource for improving and benchmarking automatic image anomaly detection algorithms which are critical for search-and-rescue missions and manhunts in complex, forested terrain. Such a dataset did not exist before. 

%-----------------------------------------------
\section*{Related Works \& Datasets}
%Lead Rakesh  (Mohamed + Nurullah help): Help on related workd (we need around 50 references, in three categories - important are datasets):
%           - image anomaly detection in the wild (color, thermal, structure) - state of the art papers and datasets
%           - image classification papers in the wild (person, animal, objects, etc) state of the art and datasets
%           - image anomaly detection not in the wild (color, thermal, structure) state of the art papers and datasets
% Differences between classifcation and anomaly detection: Classfication learns to detect know classes, Anomaly detection does not know how many / what classes exist - image anomaly detection usually pixel based decission
% our old paper on anomaly detction: https://www.mdpi.com/2673-4117/3/4/38

Anomaly detection in visual data plays a vital role in numerous real-world applications where rapid and accurate identification of irregularities is essential. Key domains include search and rescue (SAR) operations \cite{proft2015spectral,morse2012color,marshall2015color,hoai2017anomaly,cui2021dwt,agcayazi2016resquad,morandini2021anomaly,simeanomaly,seits2022evaluation,zavala2019anomaly} where identifying visual anomalies can help to locate individuals or objects of interest in vast and complex terrains, wildlife monitoring \cite{he2024anomaly} where deviations in species behavior or distribution can signal ecological threats, and automated industrial inspection \cite{rippel2021anomaly,hung2022developing,yuan2021collaborative,kahler2022anomaly,lai2018industrial,lehr2020automated,haselmann2018anomaly,tang2020anomaly,saeedi2022anomaly} where the detection of defects ensures quality control and operational safety. The nature of the task often depends on the imaging modality, leading to distinct categories such as color-based anomaly detection \cite{morse2012color,marshall2015color,hoai2017anomaly,seits2022evaluation,zavala2019anomaly,shah2023advanced,du2022anomaly}, structural anomalies (e.g., geometric deformations or missing components in manufactured parts \cite{wan2025unsupervised,bergmann2022beyond}), thermal irregularities \cite{lile2017anomaly,sledz2021thermal,lee2015model,sledz2020uav,burkard2020towards,nikolov2021seasons,tadros2024anomaly,vollmer2024detecting}, and spectral anomalies involving multispectral or hyperspectral imagery \cite{chang2002anomaly,li2018hyperspectral,jaenisch2012data,leon2021anomaly,ziemann2020temporal,su2021hyperspectral,wang2023hyperspectral,cheng2024hyperspectral,raza2022hyperspectral,xu2015anomaly,lei2019spectral}. Advances in deep learning \cite{liu2024deep,hu2022hyperspectral,singh2018deep,munir2018deepant,wang2020deep} and model-based frameworks \cite{reed1990adaptive, simon2014model,gatfaoui2022model,Mehrotra2017,xia2022gan} have significantly improved the robustness, generalization, and real-time applicability of anomaly detection systems. A variety of publicly available datasets support the development and benchmarking of anomaly detection techniques across multiple domains (see Table~\ref{tab1}). In industrial inspection, datasets such as the MVTec anomaly detection dataset (MVTec AD) \cite{bergmann2019mvtec,bergmann2021mvtec} and its successors (MVTec AD 2) \cite{heckler2025mvtec}, MVTec logical constraints anomaly detection dataset (MVTec LOCO AD) \cite{bergmann2022beyond}, MVTec 3D anomaly detection dataset (MVTec 3D-AD) \cite{bergmann2021mvtec1}) offer RGB and grayscale images of manufactured parts with various types. These datasets span resolutions from 400$\times$400 to over 4,000$\times$2,000 pixels and include thousands of samples, enabling evaluation of both 2D and 3D anomaly detection. The beanTech anomaly detection BTAD dataset \cite{mishra2021vt} provides additional variety with high-resolution RGB images. In the domain of visual surveillance, benchmark datasets such as UCSD Ped1 \cite{wang2010anomaly} and Ped2 \cite{wang2010anomaly}, CUHK Avenue \cite{lu2013abnormal}, ADOC \cite{pranav2020day}, and ShanghaiTech \cite{liu2018future} focus on anomalous motion patterns and abnormal events. These include non-pedestrian entities appearing in pedestrian walkways and unusual behaviors such as throwing objects. The datasets offer footage in either grayscale or RGB format, typically at moderate resolutions.

UAV-based anomaly detection is supported by datasets such as Drone-Anomaly \cite{jin2022anomaly}, MUAAD \cite{verma2022contextual}, and UIT-ADrone \cite{tran2023uit}, which collectively contain hundreds of thousands of RGB video frames. These datasets primarily focus on general aerial surveillance, capturing anomalous events or patterns in urban or open environments with relatively clear visibility. However, these datasets do not capture the visual complexity of forested environments, including frequent occlusions and clutter, nor are they tailored to the operational requirements of SAR and manhunt scenarios. In the thermal domain, the Thermal Anomaly Detection dataset \cite{madan2023self,nikolov2021seasons} offers nearly 40,000 infrared frames, aiding anomaly detection in low-light or thermally variable settings. For autonomous driving, datasets like RoadAnomaly21 \cite{chan2021segmentmeifyoucan}, RoadObstacle21 \cite{chan2021segmentmeifyoucan}, HTA \cite{singh2020anomalous} and FS Lost and Found \cite{blum2019fishyscapes,pinggera2016lost} provide RGB street-level imagery to identify hazardous or anomalous objects on roads. Additional benchmarks, such as VisA \cite{zou2022spot}, support large-scale industrial inspection with high-resolution RGB imagery spanning 12 objects across 3 domains, while the Street Scene \cite{ramachandra2020street} dataset captures real-world urban anomalies. In addition, several datasets using hyperspectral imagery (e.g., environmental monitoring, urban analysis, vegetation and forest assessment) \cite{jia2007spectral,zhao2014robust,gillis2002using,gader2013muufl,acito2016hyperspectral} enable applications such as spectral anomaly detection. Although these datasets collectively cover a broad spectrum of sensing modalities, resolutions, and application areas, none are designed to support anomaly detection in the highly challenging conditions of densely forested search environments.

Our dataset (Weitefeld) is the first to address search missions in complex forested terrain, such as those encountered in search-and-rescue (SAR) and manhunt operations. It is both large-scale (with tens of thousands of images and labels) and realistic, having been derived from an actual manhunt. While extensive prior work—including datasets and studies such as \cite{schedl2020search,schedl2021autonomous,song2025infrared,bondi2020birdsai,thoreau2021sarnet,sambolek2021automatic,bovzic2019deep,broyles2022wisard,bernal2024nomad,zhang2025aerial,zhang2023drone,akshatha2023manipal} supports object classification for these applications, such methods fail entirely under heavy occlusion caused by vegetation. In these scenarios, classifiers cannot detect or identify objects represented by only a few, sparsely visible pixels. Further degradation from motion blur (e.g., due to high flight speeds and low exposure) and other real-world imaging artifacts exacerbates these limitations. In contrast, anomaly detection remains robust under such conditions. However, it still relies on human intervention to classify detected objects.

\begin{table}[H]
\caption{Overview of Image Anomaly Detection Datasets}\label{tab1}%
\begin{tabular}{@{}p{3.3cm} p{1cm} p{2.5cm} p{3.2cm} p{2cm} p{3.5cm}@{}}
\toprule
Dataset & Year & Image Type & Size (Train/Val/Test) & Resolution & Application \\
\midrule
MVTec AD \cite{bergmann2019mvtec,bergmann2021mvtec} & 2019 & RGB, Grayscale & 5,354 & 700$\times$700  1,024$\times$1,024 & Industrial Inspection \\
MVTec AD 2 \cite{heckler2025mvtec} & 2025 & RGB, Grayscale & 8,004 (2,528/302/5,174) & 2,232$\times$1,024  2,448$\times$2,048  2,100$\times$1,520  4,224$\times$1,056  1,400$\times$1,900
 & Industrial Inspection\\
MVTec LOCO AD \cite{bergmann2022beyond} & 2022 & RGB & 3,644 (1,772/304/1,568) & 1,600$\times$1,280  1,600$\times$1,100  1,700$\times$1,000 1,700$\times$850  800$\times$1,600 & Industrial Inspection\\
MVTec 3D-AD \cite{bergmann2021mvtec1} & 2021 & RGB & 4,147 (2,656/294/1,197) & 800$\times$800 
400$\times$400 500$\times$500 900$\times$900 600$\times$600 900$\times$400 600$\times$800 & Industrial Inspection\\
\makecell[l]{beanTech Anomaly\\Detection (BTAD)~\cite{mishra2021vt}} & 2021 & RGB & 2,830 & 1,600$\times$1,600 600$\times$600 800$\times$600 & Industrial Inspection\\
\makecell[l]{Thermal Anomaly\\Detection~\cite{madan2023self,nikolov2021seasons}} & 2022 & Thermal & 39,600 & 288$\times$384 & Visual Surveillance\\
UCSD Ped1 \cite{wang2010anomaly} & 2010 & Grayscale & 14,000 & 238$\times$158 & Visual Surveillance\\
UCSD Ped2 \cite{wang2010anomaly} & 2010 & Grayscale & 4,560 & 360$\times$240 & Visual Surveillance\\
\makecell[l]{FS Lost and\\Found~\cite{blum2019fishyscapes,pinggera2016lost}} & 2019 & RGB & 375 & 2,048$\times$1,024 & Autonomous Driving\\
\makecell[l]{Visual Anomaly\\Dataset (VisA)~\cite{zou2022spot}} & 2022 & Grayscale & 10,821 & 4,000$\times$6,000 & Industrial Inspection\\
RoadAnomaly21 \cite{chan2021segmentmeifyoucan} & 2021 & RGB & 110 & 2,048$\times$1,024  1,280$\times$720 & Autonomous Driving\\
RoadObstacle21 \cite{chan2021segmentmeifyoucan} & 2021 & RGB & 357 & 1,920$\times$1,080 & Autonomous Driving\\
Manipal UAV Anomalous Activity
Dataset (MUAAD) \cite{verma2022contextual} & 2022 & RGB & 68,687 & 1,280$\times$720 & UAV-based Surveillance\\
Drone-Anomaly \cite{jin2022anomaly} & 2022 & RGB & 87,488 & 640$\times$640 & UAV-based Surveillance\\
CUHK Avenue \cite{lu2013abnormal} & 2013 & RGB & 30,652 & 640$\times$360 & Visual Surveillance\\
ShanghaiTech \cite{liu2018future} & 2017 & RGB & 3,17,398 & 856$\times$480 & Visual Surveillance\\
UIT-ADrone \cite{tran2023uit} & 2023 & RGB & 2,06,194 & 1,920$\times$1,080 & UAV-based Surveillance\\
HTA \cite{singh2020anomalous} & 2020 & RGB & $\sim$400,000 & 1,280$\times$720 & Visual Surveillance\\
ADOC \cite{pranav2020day} & 2020 & RGB & 2,59,127 & 2,048$\times$1,536 & Visual Surveillance\\
Street Scene \cite{ramachandra2020street} & 2020 & RGB & 2,03,257 & 1,280$\times$720 & Visual Surveillance\\
San Diego airport \cite{zhao2014robust} & 2002 & Hyperspectral & 1 & 400$\times$400$\times$224 & Remote Sensing\\
Viareggio \cite{acito2016hyperspectral} & 2013 & Hyperspectral & 1 & 450$\times$375$\times$511 & Remote Sensing\\
HYDICE Forest \cite{gillis2002using} & 1995 & Hyperspectral & 1 & 64$\times$64$\times$210 & Remote Sensing\\
HYDICE Urban \cite{jia2007spectral} & 1995 & Hyperspectral & 1 & 307$\times$307$\times$210 & Remote Sensing\\
MUUFL Gulfport (MUUFL) \cite{gader2013muufl} & 2010 & Hyperspectral & 1 & 325$\times$220$\times$64 & Remote Sensing\\
%\textbf{Weitefeld (ours)} & \textbf{2025} & \textbf{RGB} & \textbf{10,659~(labeled), 19,795~(unlabeled)}& \textbf{8,416}$\times$\textbf{6,032} & \textbf{Complex Forested Terrain (e.g., SAR, manhunt)}\\
\textbf{Weitefeld (ours)} & \textbf{2025} & \textbf{RGB} & \parbox[t]{1.5cm}{\textbf{10,659\,(labeled),\\19,795\,(unlabeled), \\ extendable}}& \textbf{8,416}$\times$\textbf{6,032} & \textbf{Complex Forested Terrain (e.g., SAR, manhunt)}\\
\bottomrule
\end{tabular}
\end{table}
%-----------------------------------------------
\section*{Methods}

The following subsection summarizes the initial data acquisition and processing steps, the online crowd search, and the subsequent data review, mapping, and ground operations that contributed to our final dataset. The Ethics Committee of Johannes Kepler University approved  the data collection and study (application number: JKU EC-55-2025). Informed consent for participation and data sharing was obtained directly from participants.  
%-----------------------------------------------
\subsection*{Data Acquisition and Processing}
\begin{figure}
\centering
\includegraphics[width=1\linewidth]{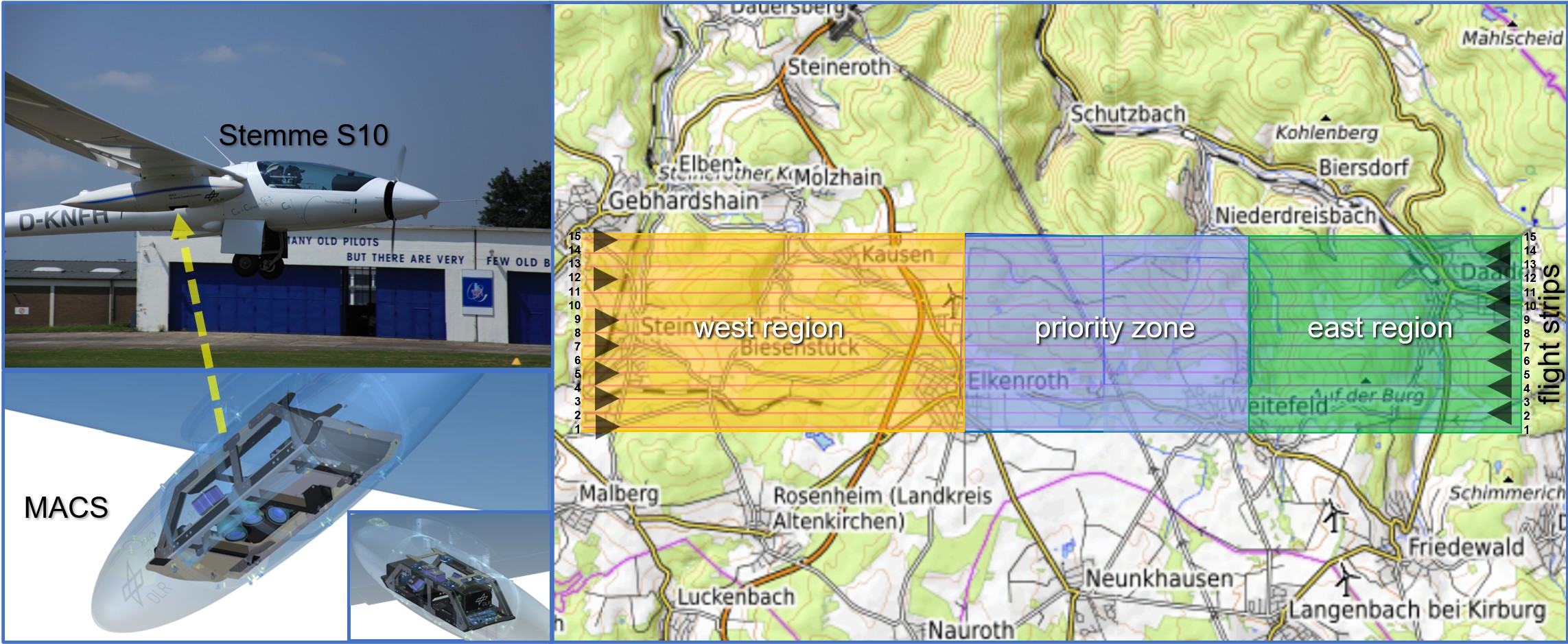}
\caption{The research aircraft (a Stemme S10 motorglider, top-left) was equipped with the Modular Aerial Camera System (MACS, bottom-left) mounted in one of its underwing pods. This system was used for data acquisition. The flight plan (right) displays the locations and numbering of the 15 flight strips. GPS coordinates for Weitefeld: 50.72468°N (latitude), 7.92714°E (longitude).}
\label{fig:data_aq}
\end{figure}

A specialized aerial camera system (MACS) developed by DLR \cite{dlr142914} mounted in an underwing pod of a Stemme S10 motorised glider was used for the image acquisition (cf. Fig. \ref{fig:data_aq}). The camera system was equipped with an industrial 50 MP RGB camera module and a 1MP Thermal Infrared sensor, which were optimized for capturing images with a high frame rate of up to 10 fps. Using the integrated dual-frequency GNSS and an industrial-grade IMU, the images were georeferenced as they were captured and stored as raw 16-bit image data. 

A flight plan was created considering the prioritized zone of interest by the police (3.5 x 2.0 km). In an altitude of 1000 ft above the highest point and with 20\% overlap across the flight direction the entire area could be mapped with 15 flight lines. The flight strips ran from south to north with an alternating east/west heading (arrows in Fig. \ref{fig:data_aq}), except for two repeated strips, and were extended to both sides to cover also a part of the low priority regions in one flight. Due to the hilly terrain in the area flown over, the resulting ground sampling distance (GSD) is between 3 and 5 cm/pixel. The flight was carried out on 27 April 2025, and to minimize shadows of occluders in the forest area, it was scheduled around the maximum solar altitude (11:26 UTC). The clear weather conditions during the image flight lead to a high radiometric scene contrast in the captured imagery. In order to focus on the dynamic range in shaded areas, assuming that the person is more likely to be there, the exposure time was set to a rather high 1.3 ms. The resulting motion blur (ground smear) and a certain overexposure in bright areas was tolerated.

After the flight the raw images were  post-processed as follows:
\begin{itemize}
    \item Removing images that mainly shown urban areas.
    \item Correction of dark signal non-uniformity, photo response non-uniformity and a color adjustment using parameters from a radiometric calibration.
    \item De-Bayering with the DCB method \cite{forsey2016demosaicing}.
    \item Gamma correction with 0.37 (adjusting tonal range by applying a power-law transformation to pixel intensity values).
    \item Color saturation was scaled by factor 2 to compensate for image degradation due to the high exposure times.
    \item Images were reduced to 24-bit color depth (8 bit per color channel) and JPEG-compressed at a rate of 60\% (13 MB typically, while keeping the full geometric resolution of 8,416x6,032 pixels) to ensure efficient online crowd searching even with limited network bandwidth.
\end{itemize}

To guide the online crowd search in such high-resolution images, we generated a binary anomaly mask for each image using the Reed-Xiaoli (RX) color anomaly detector \cite{reed1990adaptive}, with an anomaly threshold set to 0.985. The binary mask was then multiplied with the corresponding RGB image to produce a color anomaly mask. All processing was performed in parallel on 8 high-end PCs and completed in approximately 30 minutes. The resulting 10,659 image pairs (RGB + color anomaly mask) were then used for the online crowd search.
%-----------------------------------------------
\subsection*{Online Crowd Search} 
\begin{figure}
\centering
\includegraphics[width=1\linewidth]{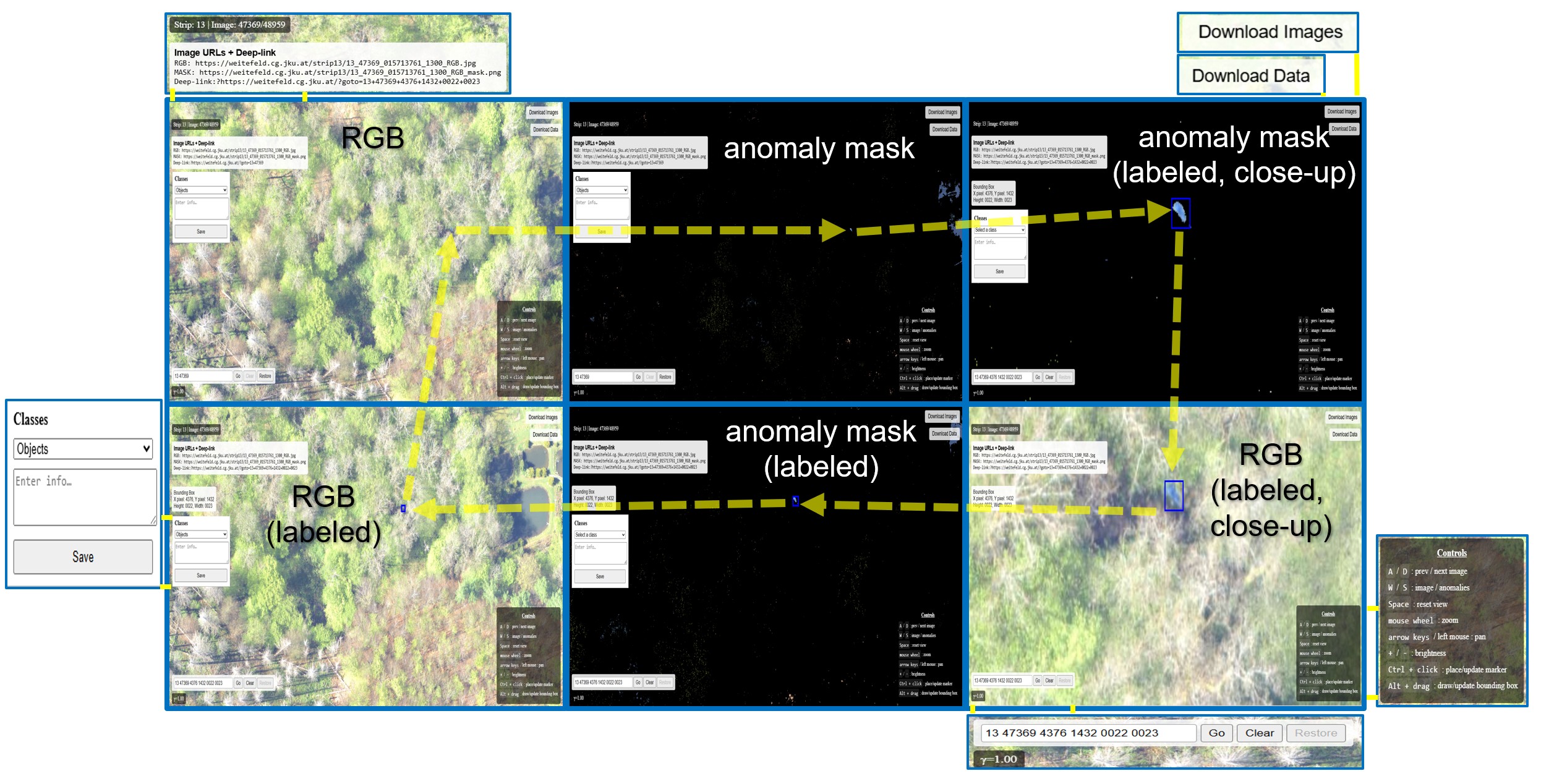}
\caption{Image analyzer web-frontend used for online crowd search and subsequent reviews of findings. It supports browsing image sequences from the same flight strip, toggling between RGB images and anomaly masks, zooming into specific areas, panning across details, adjusting brightness, and performing point- or bounding-box labeling, as well as classifying findings and adding comments.}
\label{fig:image_analyzer}
\end{figure}

For the online crowd search, an html-based and a JavaScript-based frontend (using OpenSeadragon and JSZip) has been developed (cf. Figure \ref{fig:image_analyzer}). It supports browsing through the images recorded within the same flight strip, switching between RGB images and anomaly masks, zooming and panning, brightness (gamma) changes, point (marker) and bounding box labeling, classification and commenting of findings, downloading images, and retrieving URLs and deep-links of images and findings.   

By entering a unique identifier (lower left corner), volunteers were forwarded to the first image of their assigned batch. This identifier has the following encoding: \textit{AA BBBBB}, where \textit{AA} is the flight strip number and \textit{BBBBB} is the image number. After labeling a finding, this identifier is extended as follows: \textit{AA BBBBB CCCC DDDD} if the label is a marker (point), where \textit{CCCC DDDD} are the x,y pixel coordinates of the label. \textit{AA BBBBB CCCC DDDD EEEE FFFF} if the label is a bounding box, where \textit{CCCC DDDD} are the x,y pixel coordinates of the lower left corner, and \textit{EEEE FFFF} are the height and width.     

Each volunteer was assigned overlapping batches of 110 consecutive images and instructed to examine them for anomalies that might suggest the presence of a hidden or deceased person. This included identifying conspicuous objects that seemed out of place in a forest environment, as well as potential shelters or hiding spots. To aid in their assessment, we advised volunteers to toggle between RGB images and anomaly masks, zoom in on specific areas, pan across details, and adjust brightness as needed for better visibility. Since the images in each batch were captured sequentially during flight, the gradual shift in perspective could help reveal gaps in vegetation that a single viewpoint might obscure. Participants were reminded that the precomputed anomaly masks serve only as a supplementary aid and should not be solely relied upon, as they could produce false positives or overlook relevant areas. Regions depicting villages could be disregarded. Any findings were to be marked on just one image, even if they appeared in multiple frames, and volunteers were encouraged to provide a brief subjective classification alongside each label.

The online crowd search took place from 2:00 PM on April 29th, 2025, until 8:00 AM on May 4th, 2025. During this period, a total of 405 findings were reported by 160 volunteers (members of the Rhineland-Palatinate police, students and staff of Johannes Kepler University and the German Aerospace Center, as well as students of the Berlin University of Technology).
%-----------------------------------------------
\subsection*{Data Review, Mapping, and Ground Operations}
Following the online crowd search, all findings were reviewed using the web-based frontend (cf. Figure \ref{fig:image_analyzer}). Bounding box labels were added to cases initially marked only with point labels. Of the 405 findings, 238 were flagged as relevant, while the remaining 16 --though containing anomalies-- were deemed irrelevant to the mission (e.g., cyclists on bike paths, etc.). All findings (categorized, prioritized, and annotated with comments) were then compiled into an interactive online map (cf. Figure \ref{fig:map}). Findings are represented on the map as clickable pins. Selecting a pin reveals a panel displaying key information: the finding's ID, category, volunteer-submitted comments, results from ground checks, and links to Google Maps and the Image Analyzer. Additionally, a map filter enables users to show only findings of specific categories. This map assisted police ground operations by providing precise GPS navigation to target locations via mobile devices. 

The ground operation was conducted on May 5th and 7th, 2025, during which police forces inspected all 238 relevant findings. The large-scale operation also required clearing wooded areas to access overgrown zones, as well as deploying divers and sonar scans to examine small lakes and ponds. The police reports documenting the on-site findings are included in our dataset.   

While the initial crowd searching effort labeled the 405 anomalies only in individual images, photogrammetry was used to project these labels onto every image depicting the same finding (under different perspectives and occlusion conditions). Techniques applied included bundle block adjustment, triangulation, and collinearity-based back-projection. The imaging frame rate (10 Hz) caused overlap in strip direction (up to 85 times), ultimately generating 34,424 distinct anomaly labels across 10,659 perspective images within the priority zone.

%-----------------------------------------------
\section*{Data Records}
%This section should be used to explain what the dataset contains, including the repository where it is stored, an overview of the data files and their formats, and any folder structure. Each external dataset should be cited using our data citation format. Please do not include extensive summary statistics, which should be limited to less than half a page, with 1-2 tables or figures, if required at all. Note the general expectation is that, if readers wish to scrutinise your dataset's contents, they will download and analyse it for themselves (meaning that if the figure may be generated from the data file we generally recommend it does not need to be summarised or depicted in the text). 
Our dataset is accessible through multiple sources:
\begin{itemize}  
    \item The core dataset obtained from our crowd search, used for benchmarking, is available for bulk download at: \url{https://zenodo.org/records/15848419} (approx. 144.1GB). It contains 34,424 labeled and categorized anomalies across 405 findings, captured under different viewing angles and occlusion conditions in 10,659 aerial images taken within the priority zone. 
    \item The core dataset and the additional 19,795 unlabeled aerial images covering the two non-priority zones are available for bulk download and online viewing at \url{https://weitefeld.cg.jku.at/} (approximately 404GB). The interactive web-interface (cf. Figure \ref{fig:image_analyzer}) also supports adding new labels to extend the core dataset in the future.
    \item A map overview of the entire dataset is available at \url{https://macs.dlr.de/weitefeld/} (cf. Figure \ref{fig:map}).  
\end{itemize}
%-----------------------------------------------
\subsection*{Downloading the Dataset}

 \textbf{Using the web-interface (\url{https://weitefeld.cg.jku.at/}):} Aerial images can be downloaded via the Download Images button (top-right corner) of the Image Analyzer web-frontend. To request a batch of images, users must specify the first and last image in the batch using the format \textit{AA BBBBB}, where \textit{AA} denotes the flight strip number (cf. Figure \ref{fig:data_aq}) and \textit{BBBBB} represents the image number (cf. Table \ref{tab:imagenumbers}). Upon submission, a .zip file containing the requested batch is automatically downloaded. The filenames use the following format: AA\_BBBBB\_JJJJJJJJJ\_KKKK\_RGB.jpg. In this naming convention, \textit{AA BBBBB} repeats the strip and image numbers from earlier, \textit{JJJJJJJJJ} represents a second-precise camera-internal time-stamp, while \textit{KKKK} adds microsecond-precision. A typical example would be 8\_27015\_012318083\_1300\_RGB.jpg. Note that the full image dataset has a size of 404.44GB.

\begin{table}[htbp]
  \caption{Image Numbers per Flight Strip (FIRST IMAGE → LAST IMAGE of contentious image segment).}\label{tab:imagenumbers}
      \begin{tabular}{%
      @{}%
      C{1.4cm}@{\hskip2pt}%
      C{2.3cm}@{\hskip4pt}%
      C{2.5cm}@{\hskip2pt}%
      C{2.5cm}@{\hskip2pt}%
      C{2.2cm}@{\hskip2pt}%
      C{2.5cm}@{\hskip2pt}%
      C{2.0cm}@{\hskip2pt}%
      C{1.5cm}%
      @{}%
    }
    \toprule
    \makecell{Strip\\Number}
      & \makecell{Image Numbers \\(West Region)}
      & \makecell{\#Images\\(West Region)}
      & \makecell{Image Numbers\\(Priority Zone)}
      & \makecell{\#Images\\(Priority Zone)}
      & \makecell{Image Numbers\\(East Region)}
      & \makecell{\#Images\\(East Region)}
      & \makecell{\textbf{Total}} \\
    \midrule
     01
      & 01677 → 02588 & 1,020 & 02856 → 03240 & 385 & 03521 → 03525 & 446 & \textbf{1,851} \\
      & 02748 → 02855 &     &              &     & 03545 → 03985 &      &      \\
    02
      & 05968 → 05988 & 738 & 05527 → 05967 & 441 & 04642 → 05311 & 670  & \textbf{1,849} \\
      & 06153 → 06869 &     &              &     &                &      &      \\
    03
      & 08077 → 08861 & 785 & 09067 → 09582 & 516 & 09756 → 10387 & 632  & \textbf{1,933} \\
    04
      & 12846 → 13493 & 648 & 12051 → 12721 & 671 & 11326 → 11583 & 682 & \textbf{2,001} \\
      &               &     &                &     & 11602 → 12025      &     &      \\
    05
      & 15000 → 15715 & 716 & 15865 → 16667 & 803 & 16668 → 16700 & 667 & \textbf{2,186} \\
      &               &     &                &    & 16701 → 17196 &     &      \\
      &               &     &                &    &  17262 → 17432 &     &      \\
    06
      & 19963 → 20582 & 620 & 19131 → 19838 & 708 & 18310 → 18614 & 718 & \textbf{2,046} \\
      &               &     &                &    & 18693 → 19105 &     &      \\
    07
      & 22362 → 23097 & 736 & 23133 → 23885 & 753 & 23899 → 24385 & 646 & \textbf{2,135} \\
      &               &     &                &     & 24477 → 24635 &     &      \\
    08
      & 27551 → 28214 & 664 & 26698 → 27482 & 785 & 25887 → 26129 & 712 & \textbf{2,161} \\
      &              &     &                &     & 26229 → 26697 &     &      \\
    09
      & 29664 → 30422 & 759 & 30509 → 31279 & 771 & 31280 → 31757 & 487 & \textbf{2,017} \\
      &              &     &                &     & 31886 → 31894 &     &      \\
    10
      & 34510 → 34516 & 703 & 33778 → 34509 & 732 & 33321 → 33777 & 457 & \textbf{1,892} \\
      & 34545 → 34928 &     &                &     &            &     &      \\
      & 34962 → 35273 &     &                &     &            &     &      \\
    11
      & 41176 → 41338 & 817 & 40407 → 41175 & 769 & 38666 → 38677 & 506 & \textbf{2,092} \\
      & 41382 → 41611 &     &                &     & 39913 → 40406 &     &      \\
      & 41663 → 42086 &     &                &     &                &     &      \\
    12
      & 42973 → 43499 & 865 & 44019 → 44854 & 836 & 44855 → 45321 & 467 & \textbf{2,168} \\
      & 43565 → 43727 &     &                &     &                &     &      \\
      & 43844 → 44018 &     &                &     &                &     &      \\
    13
      & 48024 → 48205 & 722 & 47264 → 48023 & 760 & 46811 → 47263 & 453 & \textbf{1,935} \\
      & 48352 → 48444 &     &                &     &                &     &      \\
      & 48513 → 48959 &     &                &     &                &     &      \\
    14
      & 57807 → 57956 & 718 & 56961 → 57806 & 846 & 56507 → 56960 & 454 & \textbf{2,018} \\
      & 58165 → 58250 &     &                &     &              &     &      \\
      & 58315 → 58796 &     &                &     &              &     &      \\
    15
      & 59901 → 59904 & 810 & 60972 → 61854 & 883 & 61855 → 62331 & 477 & \textbf{2,170} \\
      & 59919 → 59920   &     &                &     &               &     &      \\
      & 59923 → 59933   &     &                &     &               &     &      \\
      & 59936 → 60444   &     &                &     &               &     &      \\
      & 60482 → 60588   &     &                &     &               &     &      \\
      & 60795 → 60971   &     &                &     &               &     &      \\
    \textbf{Total}
      &                & \textbf{11,321} &                & \textbf{10,659} &                & \textbf{8,474} & \textbf{30,454} \\
    \bottomrule
  \end{tabular}
\end{table}

The associated findings data can be downloaded by clicking the Download Data button in the top-right corner, which provides the findings database (data.txt). Each entry follows this structured format: \textit{AA BBBBB CCCC DDDD G "HHH...HHH" AA BBBBB CCCC DDDD EEEE FFFF "III...III"}, where \textit{AA} represents the flight strip number, \textit{BBBBB} denotes the image number, \textit{CCCC DDDD} indicates the x,y pixel coordinates of a point (marker) label, \textit{G} specifies the class number (0=unknown, 1=shelter,2=object, 3=person), and \textit{HHH...HHH} contains the volunteer's comment. The second part of the entry includes repeated strip and image numbers (\textit{AA BBBBB}) followed by \textit{CCCC DDDD} representing the x,y coordinates of the lower-left corner of a bounding box label, \textit{EEEE FFFF} specifying the box's height and width, and \textit{III...III} containing the police's ground operations comment. A typical example would be \textit{07 23724 7844 0754 1 "Looks like some sort of shelter or roof of something." 07 23724 7796 0795 0074 0068 "Tarp, tent, self-made, possibly by children, near edge of town, inconspicuous."}. Missing entries are marked as \textit{NA}. For instance, an entry without a police comment appears as \textit{06 19341 0515 5798 0 "A blue thing which does not seem as it was created by nature." 06 19341 0501 5809 0026 0025 NA}. An entry containing only a point label (no bounding box or police comment) would be
\textit{07 23343 0285 1701 0 "Looks like a human-like shape with orange clothing, or wrapped in orange/bright plastic." -1 -1 -1 -1 -1 -1 "NA"}. Note, that we use -1 and "NA" to indicate integers and strings that are not available. 

 \textbf{Using Zenodo (\url{https://zenodo.org/records/15848419}):} Aerial images of the priority zone and corresponding findings from the crowd search are downloadable in 15 zip files and one .txt file. The .zip file names indicate the strip number (e.g., stip7.zip contains aerial images from flight strip 7). The naming convention of the arial images inside each .zip file is as described above. The file data.txt contains the findings database in the exact format specified above. Note that this dataset has a size of 144.1GB.    
%-----------------------------------------------
\subsection*{Viewing the Dataset Online}
The image analyzer web-frontend enables direct viewing of aerial images and associated findings in a web browser. To access specific content, users can copy one of three unique identifier formats into the text box located in the lower-left corner (see Figure \ref{fig:image_analyzer}). These identifiers, available in the findings database (data.txt), follow these patterns: The basic format \textit{AA BBBBB} displays image number \textit{BBBBB} from flight strip \textit{AA}. For point label visualization, the format extends to \textit{AA BBBBB CCCC DDDD}, where \textit{CCCC DDDD} represents the pixel coordinates within the specified image. The most comprehensive format, \textit{AA BBBBB CCCC DDDD EEEE FFFF}, displays a bounding box with its lower-left corner at \textit{CCCC DDDD} and dimensions of \textit{EEEE} (height) by \textit{FFFF} (width). Example identifiers include \textit{14 57026} for basic image display, \textit{03 09072 4497 0371} for point label viewing, and \textit{13 47790 6828 4800 0017 0016} for bounding box visualization. For detailed instructions on navigating the image analyzer web-frontend, please refer to the \textit{Online Crowd Search} section.
%-----------------------------------------------
\subsection*{Adding new Findings}
To facilitate future expansion of our dataset, the image analyzer web-frontend supports labeling of new findings. Users can annotate either additional discoveries within already-labeled aerial images of the priority zone, or in previously unlabeled images from the two non-priority zones. The interface allows creation of both point (marker) and bounding box labels. After placing a label, users can classify the finding and add comments using the controls in the top-left corner (cf. Figure \ref{fig:image_analyzer}). All annotations are subsequently saved to the findings database (data.txt). Additional guidance for using the image analyzer web-frontend is available in the \textit{Online Crowd Search} section. Note, that the core dataset obtained from our crowd search and used for benchmarking, is unchangeable. The core dataset findings occupy the first 34,424 entries in the findings database. All subsequent entries represent later additions appended to the database.  
%-----------------------------------------------
\section*{Technical Validation}
%This section should describe the experiments, analyses or checks needed to support the technical quality of the dataset, with any supporting figures and tables, as needed.
To demonstrate the applicability of our dataset, we conducted a simple benchmarking study to evaluate both deep-learning-based (Feature Reconstruction Error (FRE) \cite{ndiour2022fre}, FastFlow \cite{yu2021fastflow}, Efficient AD \cite{batzner2024efficientad}) and  model-based (Reed–Xiaoli Global (RXG) \cite{reed1990adaptive}, Reed–Xiaoli Modified (RXM) \cite{chang2002anomaly}, principal component analysis (PCA) \cite{shyu2003novel}) color anomaly detectors. Each detector operates with a predefined anomaly threshold ($t \in [0,1]$). By sweeping a range of $t$, we generated binary anomaly masks for each detector and threshold, subsequently comparing these against the annotated findings in our dataset. We assessed performance using two key metrics: \textit{Average precision}, defined as the ratio of true positives (the number of abnormal pixels correctly identified within the bounding boxes) to the sum of true positives and false positives (the total number of pixels flagged as abnormal), averaged across all images. \textit{Average detection rate}, calculated as the proportion of cases where at least one abnormal pixel was detected within the bounding box of a finding, relative to the total number of findings.

\begin{figure}
\centering
\includegraphics[width=1\linewidth]{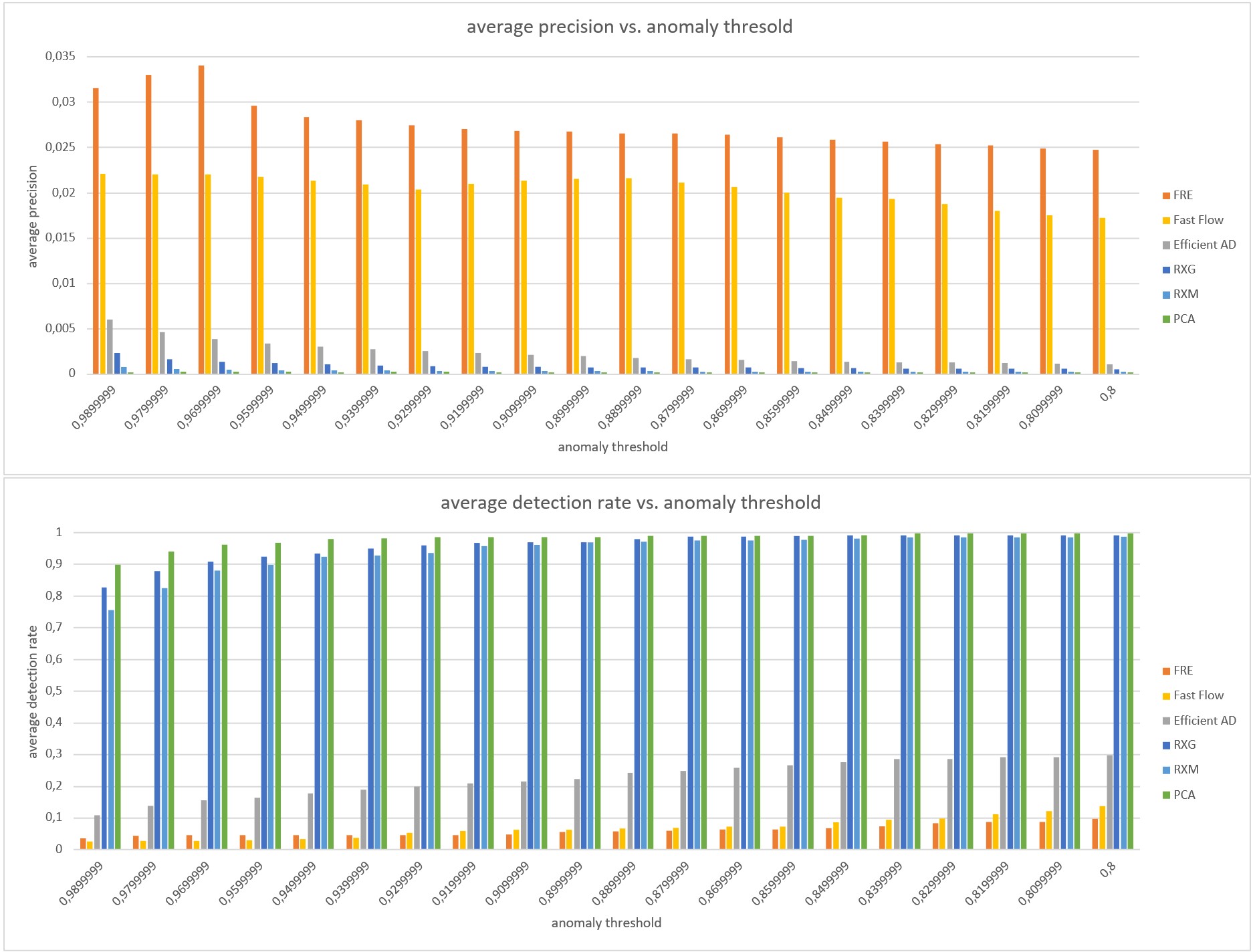}
\caption{A comparison of common deep-learning-based (FIRE, FastFlow, EfficientAD) and model-based (RX-G, RX-M, PCA) color anomaly detectors, evaluated in terms of average precision (top) and average detection rate (bottom) across a range of anomaly thresholds $t$ (x-axis).}
\label{fig:benchmark}
\end{figure}

\begin{figure}[h!]
\centering
\includegraphics[width=0.9\linewidth]{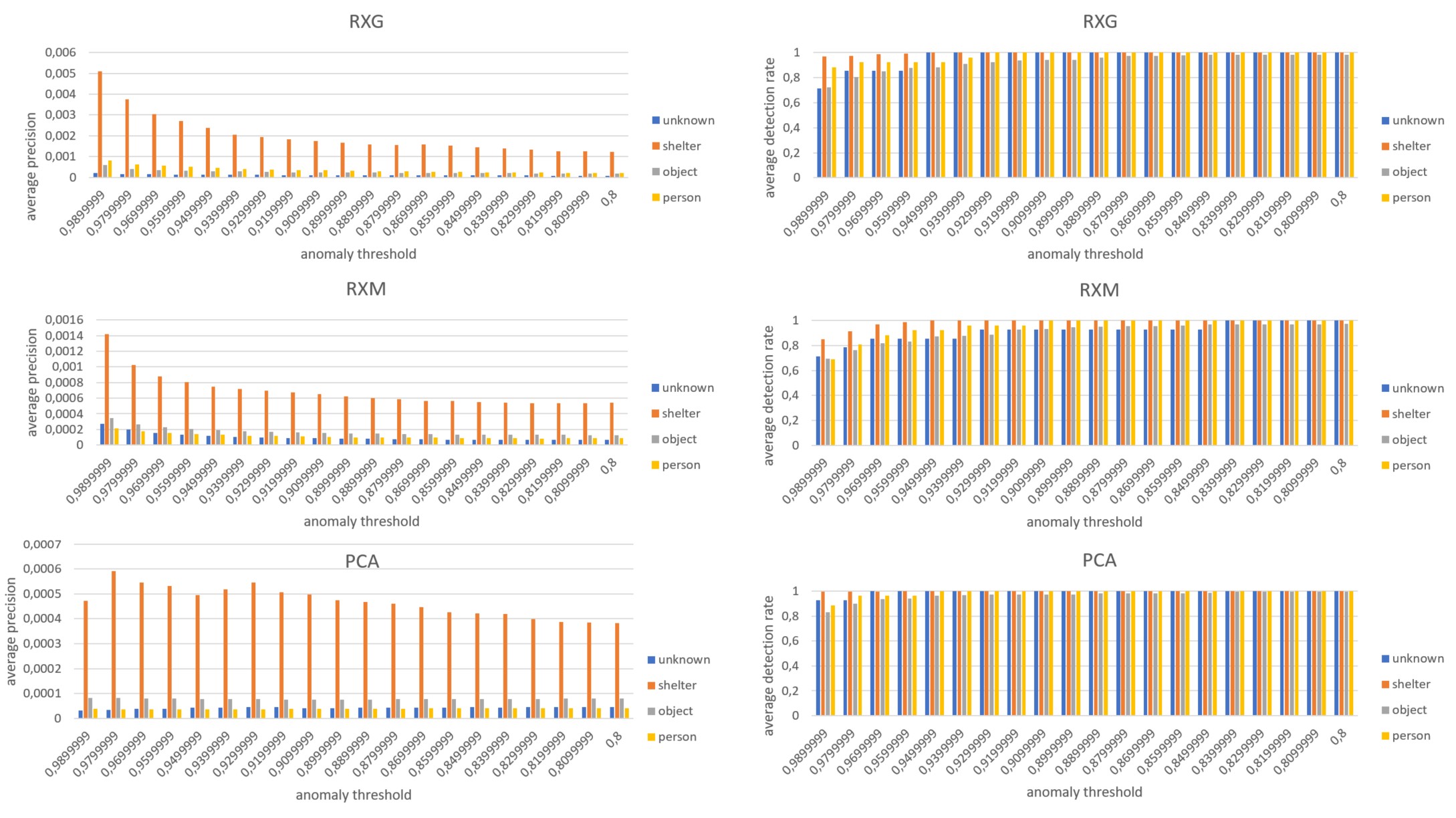}
\caption{Performance of model-based color anomaly detectors (RXG,RXM,PCA) over each class (unknown, shelter, object, person).  Anomaly threshold vs. average precision (left) and vs. average detection rate (right).}
\label{fig:benchmark_models_classes}
\end{figure}

For the deep-learning-based detectors, we employed the pre-trained models without fine-tuning. To maintain computational efficiency, we restricted our analysis to the images where findings were originally labeled by volunteers, omitting backprojections. The overall results of this evaluation are summarized in Figure \ref{fig:benchmark}. We also benchmarked the detectors for individual classes (i.e., persons, objects, shelters, and unknown). These results are presented in the Appendix (cf. Figures \ref{fig:benchmark_DL_classes} and \ref{fig:benchmark_models_classes}).
All findings, their associated bounding box annotations and classifications underwent rigorous manual verification before benchmarking to ensure label and class accuracy as far as possible. 

Overall, we observe that model-based color anomaly detectors tend to achieve low precision at high detection rates, whereas deep-learning-based approaches exhibit the opposite behavior. The former is caused by a high number of detected anomalous pixels per image (with numerous false positives), while the latter results in sparser detections (with many false negatives). We believe that this discrepancy arises because deep-learning-based models are pre-trained on ImageNet data, which contains significantly different feature representations and statistical distributions compared to our occluded anomalies. In contrast, model-based approaches rely solely on color-space statistics without accounting for higher-level image features at all.

For both types, we observe that as the anomaly threshold increases, precision decreases while the detection rate rises (this holds even when comparing methods within individual classes, as shown in Appendix Figures \ref{fig:benchmark_DL_classes} and \ref{fig:benchmark_models_classes}). A lower anomaly threshold results in more detected anomalies (both true and false). However, since false anomalies tend to occur across entire images --unlike true anomalies, which are confined to relatively small bounding boxes-- their number is significantly higher. Consequently, larger objects (e.g., shelters) generally yield higher precision values than smaller ones (see Appendix Figures \ref{fig:benchmark_DL_classes} and \ref{fig:benchmark_models_classes}).

Our experiments reveal that existing color anomaly detectors perform poorly on our dataset --both in comparison to human observers and under occlusion conditions. Deep-learning-based methods achieve precisions of below 3.5\% (or just over 10\% for individual classes) and detection rates under 30\%. Model-based approaches reach near 100\% detection rates but suffer from extremely low precision (<0.75\%). Unlike our human volunteers, these methods rely solely on local image regions or individual pixels, lacking broader contextual awareness. Anomalies detected in small regions may lose significance if they appear frequently at a global scale. These results highlight the urgent need for advancements in context-aware anomaly detection to ensure reliable performance in complex forested environments for search missions. 

It is important to note that our benchmarking study is designed solely to demonstrate the applicability of our dataset, not to evaluate the performance of individual detectors in depth. In particular, the deep-learning-based detectors could likely achieve better results if fine-tuned specifically for forested terrain.

We also evaluated automated object classification as an alternative to anomaly detection. Using our dataset, we divided the labeled samples into training, validation, and test sets and trained YOLOv12 \cite{tian2025yolov12}, a state-of-the-art classifier commonly employed in search-and-rescue missions, to detect anomalies across our four categories: persons, shelters, objects, and unknown. However, object classification completely failed with extremely low confidence scores in this experiment (0.016\% on average with a maximum of 2.6\%). The reason for this was already discussed in previous works \cite{schedl2020search,schedl2021autonomous}: Occlusion of dense vegetation obscures critical clues -- many of which occupied only a few pixels. Under such conditions, even deep neural networks struggled to generalize and automated object classification remains unrealistic.

%-----------------------------------------------
\section*{Usage Notes}
%'Usage Notes' is an optional section that can be used to provide information that may assist other researchers who reuse your data. Most commonly these are additional technical notes about how to access or process the data. Please do not use this section to write a conclusions section, general selling points, worked cases studies, or similar, as we do not publish these. 
Limitations of our dataset include the following: 1.) Anomalies are annotated using bounding boxes, as more precise segmentation masks are challenging to obtain due to partial occlusion by vegetation. 2.) The findings rely on human volunteers, introducing a degree of subjectivity in labeling. 3.) The dataset consists exclusively of aerial images captured under clear skies to maximize light reflections from densely occluded forest floors.

Note that to save new findings to the dataset, a password is required, which can be obtained from the corresponding author.   
%User ID: Weitefeld
%Password: 0i7K4nuQR9_S
%-----------------------------------------------
\section*{Code Availability}
%For all publications, a statement must be included under the subheading "Code Availability" indicating whether and how and custom code can be accessed, including any restrictions to access. This section can also include information on the versions of any software used, if relevant, and any specific variables or parameters used to generate, test, or process the current dataset if these are not included in the Methods. Please see our policy on code availability for more information. The code availability statement should be placed at the end of the manuscript, immediately before the references. 
%If no custom code has been used then the statement is still required in order to state this. 
The code for our image pre-processing and benchmark study is available at: \url{https://zenodo.org/records/15848419}.
%-----------------------------------------------
%\bibliography{}
%-----------------------------------------------
\section*{Author Contributions}
%The 'Author contributions' statement should briefly describe each author's contribution to the work. Please see also the Nature journals' authorship policies.
M.B.,M.G., and O.B. collected the data. R.J.A.A.N.,M.G.,M.M.,N.Ö., M.Y ,B.P , and O.B. processed the data. N.Ö. implemented the web interface and annotated bounding box labels. R.J.A.A.N. carried out the benchmarking study. O.B. designed, organized, carried out, and supervised the online crowd search and reviewed the data. M.M. implemented the mapping of the data. M.G. implemented the backprojection of labels. O.B., R.J.A.A.N., N.Ö., and M.G. wrote the paper. O.B., M.G., and R.B. supervised the work and managed the project. All authors reviewed the manuscript.
%-----------------------------------------------
\section*{Competing Interests}
%A 'Competing interests' statement is required for all papers accepted by and published in Scientific Data. If there is no conflict of interest, a statement declaring this must still be included in the manuscript (e.g. "The author(s) declare no competing interests"). Please see our policies for more information on what may constitute a competing interest.
The authors declare no competing interests.
%-----------------------------------------------
\section*{Acknowledgments}
%All types must include Acknowledgements, Authors contributions & Competing interest statements immediately before the References. Please note that Acknowledgements is mandated as we expect all authors to use this section to state who funded the work. 
%The 'Acknowledgements' statement should contain text acknowledging non-author contributors. Acknowledgements should be brief, and should not include thanks editors or effusive comments. Grant or contribution numbers should be included here.
We extend our sincere gratitude to all 160 volunteers who supported this mission by participating in the online crowd search. A special thanks goes to the operational command of the Rhineland-Palatinate police for their outstanding cooperation, as well as to the over 1,000 police officers in the field for their dedicated efforts.  
%-----------------------------------------------
\printbibliography %Prints bibliography
\newpage

\section*{Appendix}

\begin{figure}[h!]
\centering
\includegraphics[width=0.9\linewidth]{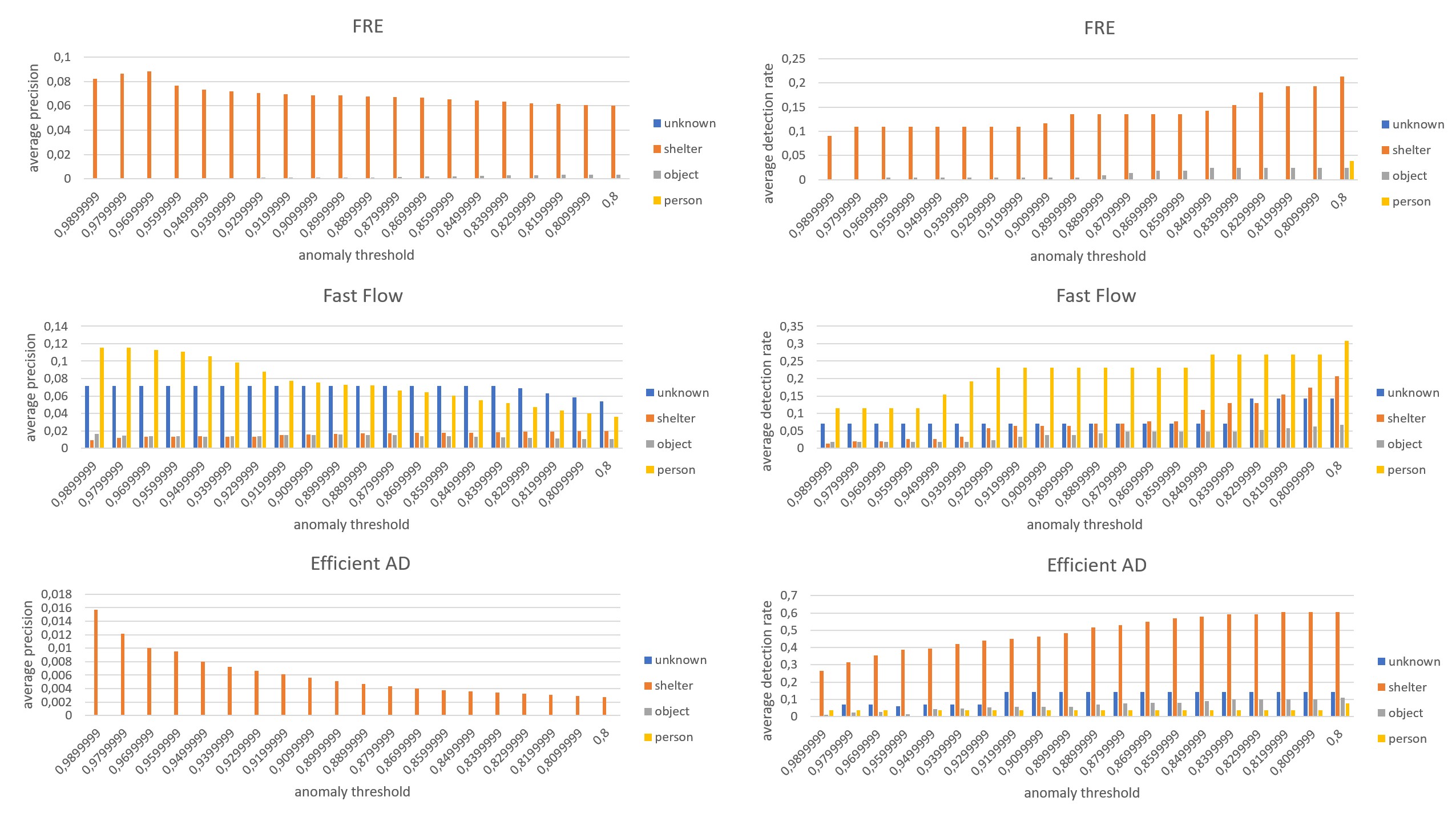}
\caption{Performance of deep-learning-based color anomaly detectors (FRE, FastFlow, EfficientAD) over each class (unknown, shelter, object, person). Anomaly threshold vs. average precision (left) and vs. average detection rate (right).}
\label{fig:benchmark_DL_classes}
\end{figure}

%-----------------------------------------------
\end{document}